# B²Net: Camouflaged Object Detection via Boundary Aware and Boundary Fusion

Junmin Cai, Han Sun*, Ningzhong Liu

*College of Computer Science and Technology
Nanjing University of Aeronautics and Astronautics, Nanjing 210016, China
Email: sunhan@nuaa.edu.cn

*Abstract*

**Camouflaged object detection (COD) aims to identify objects in images that are well hidden in the environment due to their high similarity to the background in terms of texture and color. However, existing most boundary-guided camouflage object detection algorithms tend to generate object boundaries early in the network, and inaccurate edge priors often introduce noises in object detection. Address on this issue, we propose a novel network named B²Net aiming to enhance the accuracy of obtained boundaries by reusing boundary-aware modules at different stages of the network. Specifically, we present a Residual Feature Enhanced Module (RFEM) with the goal of integrating more discriminative feature representations to enhance detection accuracy and reliability. After that, the Boundary Aware Module (BAM) is introduced to explore edge cues twice by integrating spatial information from low-level features and semantic information from high-level features. Finally, we design the Cross-scale Boundary Fusion Module(CBFM) that integrate information across different scales in a top-down manner, merging boundary features with object features to obtain a comprehensive feature representation incorporating boundary information. Extensive experimental results on three challenging benchmark datasets demonstrate that our proposed method B²Net outperforms 15 state-of-art methods under widely used evaluation metrics. Code will be made publicly available.**

## I. INTRODUCTION

Camouflage is a survival skill that animals acquire with constant evolution. In order to avoid predators or prey on other animals, animals employ several camouflage skills to make themselves difficult to be identified. Camouflage Object Detection (COD) [1] is an emerging and challenging task, which aims to search and segment camouflaged objects from single image. With its rich history in biology, art, and the military [2], COD also has significant application value in various fields such as locust detection, polyp segmentation [3], marine animal segmentation [4], and recreational art [5]. However, due to the nature of camouflage, characterized by the high intrinsic similarities between candidate objects and chaotic backgrounds, COD presents greater challenges compared to general object detection tasks.

To tackle this issue, numerous deep learning-based methods have been proposed for camouflaged object detection and have shown great potentials. Fan et al. [6] constructed the COD10K dataset which contains 5066 samples of camouflaged objects and proposed a search-identification network (SINet). SINet first uses a search module to roughly locate the camouflaged object, and then uses an identification module for precise segment. Inspired by the design principles of SINet, a variety of approaches focusing on cross-level feature fusion have been proposed [7], [8]. Although these models have improved camouflaged object detection from a local perspective, they still cannot obtain clear boundaries. In COD, the high similarity between camouflaged objects and their surroundings makes boundary information between the object and the background particularly important. Therefore, the extraction of boundary information is still a key factor.

To overcome the above limitation, researchers have introduced edge cues to enhance the representation of segmentation features [9] [10]. The common practice is to generate an edge prediction at early stage of the network and use it as edge prior to guide the fusion of multi-level segmentation features. However, these methods still suffer from two shortcomings: (1) low-level features that preserve low-frequency texture information are less focused. (2) the importance of the accuracy of the initially generated edge predictions is often overlooked in favor of prioritizing the incorporation of edge priors into segmentation features. Due to the lack of semantics information, edge prediction generated early is prone to lose the integrity of the camouflaged object, which misleads segmentation into erroneous foreground prediction.

To this end, in this paper, we propose a novel network(B²Net) via boundary aware and cross-scale boundary fusion, which explicitly employs edge semantic to enhance the performance of camouflaged object detection. The improved Pyramid Vision Transformer (PVTv2) [11] is adopted as the backbone to extract global contextual information at multiple scales effectively. Then, a residual feature enhanced module(RFEM) is designed to refine the features at each scale. Further, we design a simple yet effective boundary aware module (BAM), which integrates the low-level local spatial information and high-level global location information to explore edge semantics related to object boundaries under explicitly boundary supervision. Finally, the cross-scale boundary fusion module (CBFM) is introduced to aggregate the edge features with the camouflaged object features gradually from top to bottom to guide the representation learning of COD. Benefiting from the well-designed modules, the proposed B$^2$Net predicts camouflaged objects with fine object structure and boundaries.

To sum up, our main contributions are as follows:
1) For the COD task, we propose a novel boundary-guided network, i.e., B$^2$Net, which excavates and integrates boundary-related edge semantics to boost the performance of camouflaged object detection.
2) We carefully design the boundary aware module (BAM) and cross-scale boundary fusion feature module (CBFM) to enhance boundary semantics and explore valuable and powerful feature representation for COD.
3) B$^2$Net is validated on three popular COD datasets and achieves the most outstanding performance among 15 COD methods, especially when the camouflaged objects have very blurred boundaries and similar colors/patterns with their backgrounds.

II. RELATED WORK

*A. Bio-inspired methods*

Recently, bio-inspired methods utilizing deep learning have emerged as the dominant approach in the field of COD. Motivated by how animals hunt in the wild, Fan et al. [6] designs a search module and an identification module to detect camouflaged objects by simulating the process of animal hunting.

By exploiting the sensory and cognitive mechanisms towards initial detection and predator learning of predation process, Zhang et al. [12] propose a PreyNet, which mimics the two processes of predation, leads to better understanding and significant performance improvement for camouflaged object detection.

Based on the observation that people usually search for camouflaged objects by zooming and out, Pang et al. [13] propose a mixed-scale triplet network to better combines information between multi-scale features explicitly.

*B. Edge-guidance methods*

In addition to bio-inspired methods, an increasing number of approaches have integrated boundary cues as a means to improve performance.

To further clarify the indistinct boundaries, Sun et al. [9] designs an edge-guidance feature module to embed the edge cues into segmentation features. Zhu et al. [10] utilizes adaptive space normalization to do this in a more effect way. And in order to ensure the accuracy of the generated boundary, Sun et al. [14] designed an edge-aware mirror network which models edge detection and camouflaged object segmentation as a cross refinement process, accompanied by an increase in the complexity of the model.

Different from the existing methods, our method focuses on boundary representation by simply reusing the boundary aware module, without significantly increasing the complexity of the model, thereby improving the accuracy of the model.

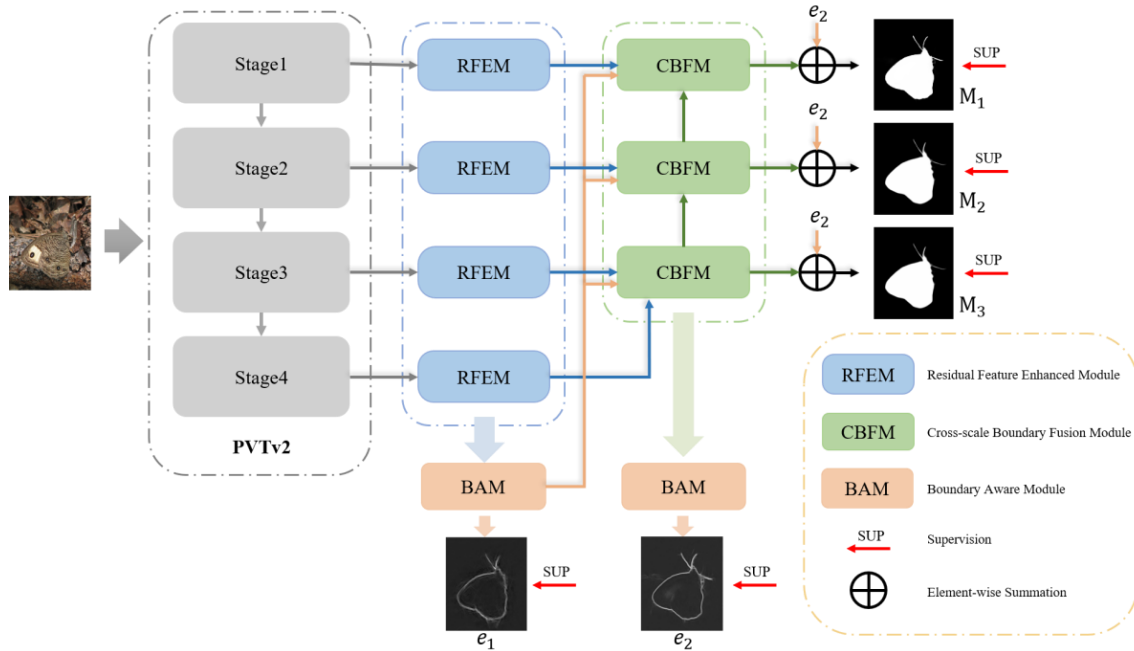

Fig. 1. The overall architecture of the proposed B$^2$Net, which consists of three key components, i.e., Residual Feature Enhanced Module (RFEM), Boundary Aware Module(BAM) and Cross-scale Boundary Fusion Module(CBFM).

## III. OUR METHOD

### A. Overview

Fig. 1. shows the overview of the proposed B²Net, which consists of three kinds of key components including Residual Feature Enhanced Module, Boundary Aware Module, and Cross-scale Boundary Fusion Module.

Specifically, given an input image I, we first adopt the PVTv2 [11] as backbone to extract features at four levels, which can be denoted as $F = \{f_i, i = 1,2,3,4\}$. Then, we feed $F$ into Residual Feature Enhanced Module to extract multi-level edge features with the channel size of 64 denoted as $F' = \{f'_i, i = 1,2,3,4\}$. Then, a boundary aware module (BAM) is applied to excavate initiatory object-related edge semantics ($e1$) from the low-level features, which contain local edge details ($f1, f2$), and the high-level features, which contain global location information ($f4$) under object boundary supervision. After that, multiple Cross-scale Boundary Fusion Module(CBFM) are employed to progressively integrate the edge cues from BAM with multi-level features at each level to guide feature learning in a top-down manner and discover camouflaged objects. Moreover, the output of CBFM will be fed into a new BAM module to obtain more accurate edge semantics ($e2$). Finally, the $e2$ will be concatenated with the output of CBFM and generate the predict map $M_i, i \in \{1,2,3\}$, and we will adopt the $M_1$ as the final predict map, which serves as the output of the entire network. Multiple side-out supervised strategies are implemented to boost the COD performance. We will provide the details of each key component below.

### B. Residual Feature Enhanced Module

To further enrich the contextual information obtained by PVTv2 at various scales, the Residual Feature Enhanced Module(RFEM) it designed with inspiration from the Inception module and Res2Net block [15]. As shown in Fig. 2, RFEM adopts the 3×3 convolution in parallel while employing residual blocks to enlarge the receptive fields successively. To be more specific, for an input feature $f_i$, we utilize 4 branches to capture different characterizations. Each branch is equipped with a 1×1 convolution to reduce the number of channels, a 1×3 and 3×1 asymmetric convolution for reducing the computational load. The output of each branch is added to the input of the next branch. The general formulation of the operation is defined as

$$Bout_k^i = \begin{cases} Conv_r(f_i) & k = 1 \\ Conv_r(f_i \oplus Bout_{k-1}) & k = 2,3,4 \end{cases} \quad (1)$$

where $f_i$ denotes the $ith$ feature map produced by the backbone network, $k$ is the branch number, $Bout_k^i$ denotes the output of the $kth$ branch, $\oplus$ is element-wise addition, $Conv_r()$ denotes the stacked convolutional layer mentioned above. After that, we concatenate the outputs of 4 branches followed by a 1×1 convolution to adjust the channel to 64 and add it to the input feature $f_i$. Finally, we obtain the output feature $F' = \{f'_i, i = 1,2,3,4\}$ embedded with multi scale information which is computed as

$$f'_i = Conv(f_i) \oplus Conv(Cat_{k=1}^4(Bout_k^i)), \quad (2)$$

where $Conv()$ denotes 1×1 convolution, $Cat_k^4 = 1$ denotes the concatenation of all 4 branches.

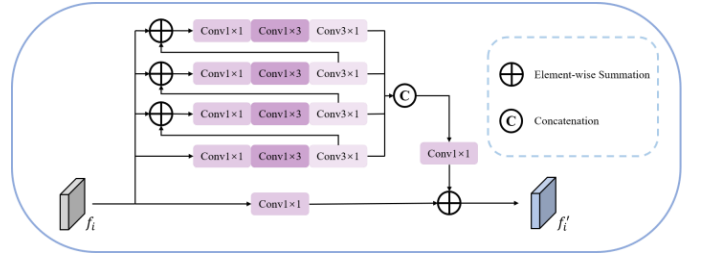

Fig. 2. The detailed architecture of the proposed Residual Feature Enhanced Module(RFEM).

### C. Boundary Aware Module

Boundary information is of great significance for the segmentation and localization of camouflaged objects. Camouflage images have high resolution in low-level feature regions and can extract boundary information, but they are inevitably interfered with by the boundary information of non-camouflaged objects in the background, resulting in rough image boundaries. To this end, we propose a boundary aware module (BAM) to extract boundary features related to camouflaged objects.

The specific structure of BAM is shown in Fig. 3. Take the BAM that accepts the REFM as input in Fig. 1. as an example, we first add $f'_1$ and $f'_2$. to represent the initial feature $f'_{12}$, and then we multiply the semantic location information $f'_4$ and $f'_{12}$ to locate the camouflaged object to avoid interference from other irrelevant features, and then perform 3×3 convolution and skip connections to obtain spatial features with rich spatial information and accurate localization. Finally, we use spatial attention [16] to enhance the feature representation. The following formulas can describe the whole process above:

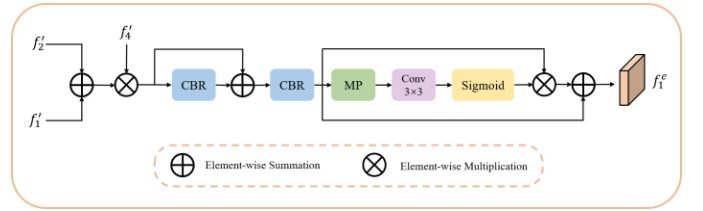

Fig. 3. The detailed architecture of the proposed Boundary Aware Module(BAM).

$$\begin{cases} f'_{124} = f'_4 \otimes (f'_1 \oplus f'_2) \\ f''_{124} = F_{3\times3}(F_{3\times3}(f'_{124}) \oplus f'_{124}) \\ f^e = f''_{124} \otimes \sigma(Cov_{3\times3}(MP(f''_{124}))) \oplus f''_{124} \end{cases} \quad (3)$$

where $F_{3\times3}$ represents 3×3 convolution followed by batch normalization and ReLU activation function. $MP(\cdot)$ denotes max pooling. $\sigma(\cdot)$ is sigmoid function.

It is worth noting that the BAM module is executed twice, the second time taking the output of the aforementioned CBFM as input to the module to produce a more accurate boundary result, as shown in Fig. 1.

### D. Cross-scale Boundary Fusion Module

The Cross-scale Boundary Fusion Module(CBFM) is designed to inject boundary-related edge cues into the representation learning to enhance the feature representation with object structure semantics. And it is known that the scale variation [17] is a crucial issue in object detection since the scales of objects may vary significantly across different scenarios. The low-level features at fine scales are important to the detection of small objects, while the high-level features at coarse scales are more sensitive to large objects. Therefore, the diversity of scale information is of great significance to the detection accuracy. In our approach, we combine CBFM in a top-down manner, progressively aggregating multiple levels of fusion features and discovering camouflaged objects.

As shown in Fig. 4, given the input shallow feature $f'_i, i \in \{1,2,3\}$ and the edge feature $f^e$, we first perform the element-wise multiplication between them with a 3×3 convolution and a learnable parameter $\alpha$ to obtain the initial fused features $f^{if}_i$, which can be denoted as:

$$f^{if}_i = \alpha \cdot (F_{3\times3}(f'_i \otimes f^e)) \quad (4)$$

where $\alpha$ is a learnable parameter. $F_{3\times3}$ represents 3×3 convolution followed by batch normalization and ReLU activation function.

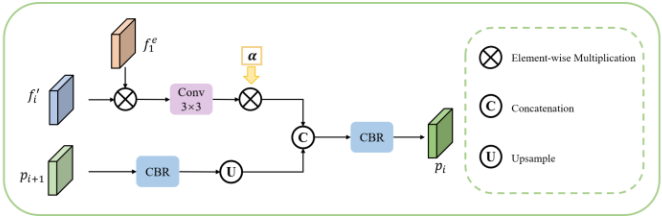

Fig. 4. The detailed architecture of the proposed Cross-scale Boundary Fusion Module(CBFM).

Then we use sequential operations which consist of a 3×3 convolution layer, batch normalization, ReLU activation function, and bilinear upsampling operation on $p'_{i+1}$ to align their shapes and features and "Concat" them to get fused features. And the sequential operations are also utilized to refine the fused features and obtain the output $p_i$, the whole process can be formulated as follows:

$$p_i = F_{3\times3}(Cat(UP(F_{3\times3}(p_{i+1})), f^{if}_i)) \quad (5)$$

where $p_{i+1}$ is the output of $CBFM_{i+1}$, $F_{3\times3}$ represents 3×3 convolution followed by batch normalization and ReLU activation function, $Cat(\cdot)$ represents concatenation function with dim = 1.

### E. Loss Function

We employ a combination of loss functions to supervise the output map in our model: Weighted Intersection Over Union $L^{IOU}_\omega$ and weighted Binary Cross-entropy $L^{BCE}_\omega$. IOU in image segmentation is generally responsible for maintaining the structure of the predicted map following the ground truth. Adding a weighting factor to the IOU loss enhances its performance by assigning different weights to the class regions based on their importance in the task. On the other hand, Binary Cross-entropy is the pixel-wise classification, and $L^{BCE}_\omega$ similarly performs better on complex samples. Combining these losses enables the network to focus more on complex samples, which are expected to rise in the COD.

The $L^{IOU}_\omega$ and $L^{BCE}_\omega$ losses are computed on the predictions of three camouflaged object masks $M_i, i \in \{1,2,3\}$ Similarly, to supervise the edge ($e_i, i \in \{1,2\}$), we use Dice Loss $L^{Dice}$ to improve the edge quality. By incorporating these supervisory signals at different stages into the training process, our proposed model can effectively learn to segment the hidden object in complex scenes. Eq. 6 provides the total loss $L_{total}$ for the network's supervision.

$$L_{total} = \sum_{i=1}^{3}(L^{BCE}_w(M_i, G_m) + L^{IOU}_w(M_i, G_m)) + \sum_{i=1}^{2} L^{Dice}(e_i, G_e)). \quad (6)$$

## IV. EXPERIMENT

### A. Implementation Details

The proposed approach is implemented using PyTorch. The Adam optimizer is employed to update the network parameters. The input images are resized to 352 × 352 pixels using the bilinear interpolation. The initial learning rate is set to 8e-5 and a weight decay of 0.1 is used in network training. The batch size is set to 16 and the number of maximum epochs is set to 100. It takes 6 hours to about accomplish the training process on an NVIDIA Titan Xp GPU.

### B. Datasets and Evaluation Metrics

TABLE I
FOUR EVALUATION METRICS ARE EMPLOYED IN THIS STUDY, NAMELY $S_\alpha \uparrow$ $F_\beta^\omega \uparrow$, $E_\phi \uparrow$, AND $M \downarrow$. THE SYMBOLS "↑" AND "↓" INDICATE THAT LARGER AND SMALLER VALUES ARE BETTER, RESPECTIVELY. THE BEST RESULTS ARE HIGHLIGHTED IN BOLD.

| Method | Year | COD10K-Test | | | | Camo-Test | | | | NC4K-Test | | | |
|---|---|---|---|---|---|---|---|---|---|---|---|---|---|
| | | $S_\alpha$ | $E_\phi$ | $F_\beta^\omega$ | M | $S_\alpha$ | $E_\phi$ | $F_\beta^\omega$ | M | $S_\alpha$ | $E_\phi$ | $F_\beta^\omega$ | M |
| SINet | 2020 | 0.776 | 0.864 | 0.631 | 0.043 | 0.745 | 0.829 | 0.644 | 0.092 | 0.808 | 0.871 | 0.723 | 0.058 |
| PraNet | 2020 | 0.789 | 0.861 | 0.629 | 0.045 | 0.769 | 0.837 | 0.663 | 0.094 | 0.822 | 0.876 | 0.724 | 0.059 |
| C2FNet | 2021 | 0.813 | 0.890 | 0.686 | 0.036 | 0.796 | 0.864 | 0.719 | 0.080 | 0.838 | 0.897 | 0.762 | 0.049 |
| PFNet | 2021 | 0.800 | 0.877 | 0.660 | 0.040 | 0.782 | 0.842 | 0.695 | 0.085 | 0.829 | 0.887 | 0.745 | 0.053 |
| SINet-V2 | 2022 | 0.815 | 0.887 | 0.680 | 0.037 | 0.820 | 0.882 | 0.743 | 0.070 | 0.847 | 0.903 | 0.770 | 0.048 |
| ZoomNet | 2022 | 0.838 | 0.911 | 0.729 | 0.029 | 0.82 | 0.892 | 0.752 | 0.066 | 0.853 | 0.912 | 0.784 | 0.043 |
| BSANet | 2022 | 0.817 | 0.887 | 0.696 | 0.035 | 0.804 | 0.860 | 0.728 | 0.079 | 0.842 | 0.897 | 0.771 | 0.048 |
| BGNet | 2022 | 0.831 | 0.901 | 0.722 | 0.033 | 0.816 | 0.871 | 0.751 | 0.069 | 0.851 | 0.907 | 0.788 | 0.044 |
| EAMNet | 2023 | 0.839 | 0.907 | 0.733 | 0.029 | 0.831 | 0.890 | 0.763 | 0.064 | 0.862 | 0.916 | 0.801 | 0.040 |
| FDNet | 2023 | 0.857 | 0.918 | 0.763 | 0.028 | 0.836 | 0.886 | 0.777 | 0.066 | 0.865 | 0.911 | 0.803 | 0.042 |
| FEDER | 2023 | 0.844 | 0.911 | 0.748 | 0.029 | 0.802 | 0.873 | 0.738 | 0.071 | 0.847 | 0.915 | 0.789 | 0.044 |
| DGNet | 2023 | 0.822 | 0.911 | 0.692 | 0.033 | 0.838 | 0.914 | 0.768 | 0.057 | 0.857 | 0.922 | 0.783 | 0.042 |
| FSPNet | 2023 | 0.851 | 0.930 | 0.735 | 0.026 | 0.856 | 0.928 | 0.799 | 0.05 | 0.878 | 0.937 | 0.816 | 0.035 |
| CRINet | 2024 | 0.819 | 0.903 | 0.733 | 0.035 | 0.810 | 0.883 | 0.791 | 0.072 | 0.848 | 0.911 | 0.811 | 0.046 |
| LGPNet | 2024 | 0.832 | 0.899 | 0.763 | 0.032 | 0.817 | 0.873 | 0.797 | 0.070 | 0.856 | 0.908 | 0.827 | 0.044 |
| **Ours** | **-** | **0.862** | **0.934** | **0.772** | **0.023** | **0.866** | **0.935** | **0.808** | **0.048** | **0.882** | **0.945** | **0.829** | **0.033** |

We conducted experiments on four publicly available camouflage object detection benchmark datasets: CAMO [18], CHAMELEON [19], COD10K [6], and NC4K [20]. In this paper, the training sets of CAMO, and COD10K as the model's training set, with a total of 4,040 images, and other data as the model's test set.

To present an in-depth evaluation of the performance of COD algorithms, we use five evaluation metrics that are widely used in image segmentation, including Enhanced-measure [21], Structure-measure [22], weighted F-measure [23] and Mean Absolute Error [24] denoted as $S_\alpha$, $E_\phi$, $F_\beta^\omega$ and M, respectively.

*C. Comparison with State-of-the-Art Methods*

We compared our proposed method with 15 state-of-the-art models, including SINet [6], PraNet [3], C2FNet [25], PFNet[26], SINet-v2 [1], ZoomNet [3], BSANet [10], BGNet[9], EAMNet[14], and FDNet [27], FEDER [28], DGNet [29], FSPNet [30], CRINet [31], LGPNet [32].

Quantitative Comparison: We exploit all four metrics to compare with SOTAs. Table I summarizes the results of all methods on three benchmark datasets, where the best ones are highlighted in bold. It is obvious that our method outperforms all other models on three datasets under four evaluation metrics. In particular, while compared with the second-best FSPNet, our method increases $S_\alpha$ by 0.93%, $E_\phi$ by 0.68% and $F_\beta^\omega$ by 2.5% on average. Compared with the third-best FDNet, our method increases $S_\alpha$ by 2.06%, $E_\phi$ by 3.67% and $F_\beta^\omega$ by 2.67% on average. Overall, our model achieves the best performance based on the available performance statistics and analysis.

Besides, BGNet, BSANet, and EAMNet utilize auxiliary edge or boundary information and still fail to locate camouflaged objects, while our model can effectively locate them and achieve the best performance. This is because our proposed boundary aware module(BAM) and cross-scale boundary fusion module(CBFM), which jointly operate on ambiguous regions, explore effective representation of edge features, significantly improving the performance of COD.

Visual Comparison: From the visual results in Fig. 5, our model outperforms the comparative models in the visual comparison results of the selected collected test samples. Specifically, in the first and second row, the camouflaged objects have similar textures to the background, which poses a severe challenge for identifying them from similar backgrounds. In this case, our model performs better, accurately locating the camouflaged object. From the third to the fifth rows, it can be observed that our model demonstrates certain advantages in detecting elongated objects. Not only in the identification of camouflaged object areas, but also in the quality of generated boundaries, it surpasses other models. And in the sixth and seven row, when camouflaged objects are partially occluded, our model can accurately identify multiple segments of camouflaged objects and generate relatively accurate boundaries. Overall, the results demonstrate that our predictions have clearer and more complete object regions, as well as sharper contours, and our model can perform well in detecting camouflaged objects under different challenging scenes.

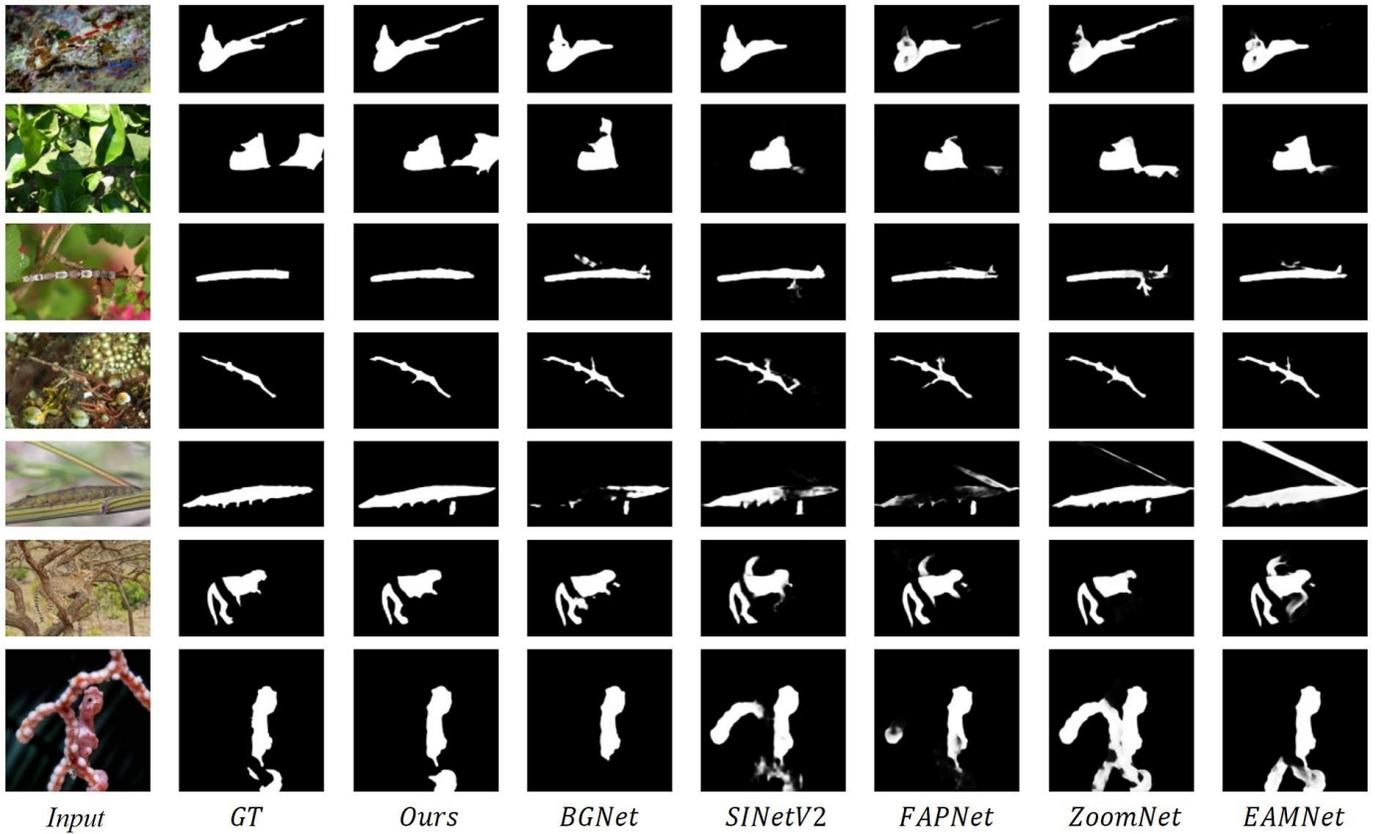

Fig. 5. This section provides a visual comparison between our proposed model and the latest models. Specifically, we show (a) the input image, (b) the ground truth (GT), (c) our method, and (d)–(h) the state-of-the-art models, including BGNet, SINetV2, FAPNet, ZoomNet, and EAMNet. The results demonstrate that our model achieves superior performance to the other compared methods in terms of visual quality.

Failure Cases: We present a number of typical failure cases in Fig. 6. As can be observed, our model may fail in some extremely challenging cases, such as tiny object scenarios. This may be because it is more difficult to obtain the object boundary in tiny object scenarios. However, it is worth noting that in these cases, existing state-of-the-art (SOTA) models also face challenges. This indicates that the issue in tiny object scenarios is a difficult problem that researchers in the field are still striving to solve, and there is not yet a widely accepted optimal solution. Nevertheless, our model demonstrates relatively superior performance in these highly challenging scenarios, further validating the effectiveness and potential of our approach.

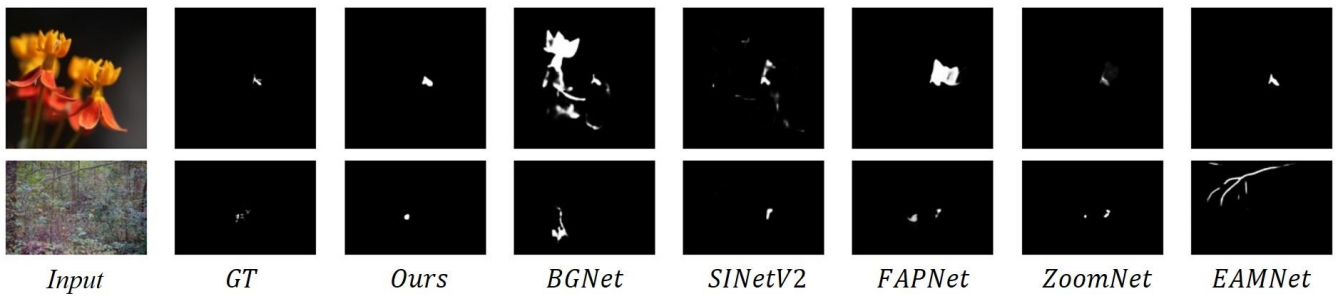

Fig. 6. Some failure cases of our model and five comparison methods.

TABLE II
QUANTITATIVE EVALUATION FOR ABLATION STUDIES ON THREE DATASETS. THE BEST
RESULTS ARE HIGHLIGHTED IN BOLD.

| Method | COD10K-Test | | | | Camo-Test | | | | NC4K-Test | | | |
|---|---|---|---|---|---|---|---|---|---|---|---|---|
| | $S_\alpha$ | $E_\phi$ | $F_\beta^\omega$ | M | $S_\alpha$ | $E_\phi$ | $F_\beta^\omega$ | M | $S_\alpha$ | $E_\phi$ | $F_\beta^\omega$ | M |
| M1 | 0.805 | 0.872 | 0.711 | 0.040 | 0.802 | 0.866 | 0.745 | 0.072 | 0.812 | 0.872 | 0.766 | 0.053 |
| M2 | 0.839 | 0.902 | 0.752 | 0.030 | 0.831 | 0.899 | 0.773 | 0.061 | 0.839 | 0.896 | 0.795 | 0.044 |
| M3 | 0.841 | 0.905 | 0.751 | 0.028 | 0.838 | 0.906 | 0.779 | 0.059 | 0.848 | 0.913 | 0.801 | 0.041 |
| M4 | 0.857 | 0.927 | 0.766 | 0.025 | 0.853 | 0.931 | 0.798 | 0.052 | 0.871 | 0.932 | 0.822 | 0.035 |
| **M5(Ours)** | **0.862** | **0.934** | **0.772** | **0.023** | **0.866** | **0.935** | **0.808** | **0.048** | **0.882** | **0.945** | **0.829** | **0.033** |

*D. Ablation Study*

We conduct a set of ablation experiments to verify the effectiveness of the BAM module, the CBFM module and out proposed new architecture in our B$^2$Net. Specifically, the following five models are mainly involved in our ablation study:

1) The Baseline model is composed of the backbone PVTv2 and residual feature enhanced module(RFEM).
2) Baseline + BAM (M2): Adding the BAM modules into the Baseline (M1).
3) Baseline + CBFM (M3): Adding the CBFM modules into the Baseline (M1). The CBFM module in this case removes the relevant part of edge feature fusion and only takes the features between adjacent layers as input.
4) Baseline + CBFM + 1BAM (M4): Adding one BAM modules into the Baseline + CBFM (M3). The output of the only BAM module is also the input to the CBFM.
5) Baseline + CBFM + 2BAM (M5): The completed model.

Table II lists the quantitative results of different models in the ablation study on three benchmark datasets.

Effectiveness of BAM: From M2 and M1 in the Table II, it can be seen that the adding BAM module could improve overall performance effectively. This owes to BAM's ability to exploit rich spatial fine-grained information from low-level features. Thus, the edge prior extracted by BAM is beneficial to boost detection performance.

Effectiveness of CBFM: This module helps the model extract valid information from the edge feature, thus making the segmentation results more accurate. Compare M2 and M5 in Table II, $F_\beta^\omega$ increase by 5.63% in COD10K, and 4.57% in NC4K. This fully demonstrates the effectiveness of combining CBFM modules in a top-down manner to fuse edge features. Besides, from M1 and M3 in the Table II, it can be seen that the CBFM module itself has a good enhancement effect on features.

Effectiveness of Our Proposed New Architecture: Most of the previous COD methods introduce the boundary tend to generate the boundary results at the early stage of the network, but our method reuses the proposed BAM module after feature fusion to obtain more accurate boundary results.

To further verify the effectiveness of our proposed network, we have conducted experiments where we integrated our boundary reuse strategy into other existing methods, as can be seen in Table 3. These experiments were performed using the same backbone network to provide a direct comparison. The results of these experiments clearly demonstrating the effectiveness of our proposed network. This not only validates our approach but also highlights the improvements achieved by incorporating our boundary reuse strategy.

It also can be seen from M4 and M5 in Table III that the new architecture proposed by us simply and effectively improves the detection performance. Moreover, as can be seen from Fig. 7, the boundary result obtained by the remultiplexed BAM is indeed improved to some extent compared with the boundary graph generated for the first time, which also shows the effectiveness of the CBFM module.

V. CONCLUSION

In this paper, we propose a novel framework via boundary aware and cross-scale boundary fusion, named B$^2$Net. In particular, The BAM module generates the boundary twice by fusing the spatial information contained in the low-level feature and the semantic information contained in the high-level feature. The cascaded CBFM combines edge features and enhancement features acquired in a top-down manner, facilitating the fusion of information across different scales. Extensive experiments on three benchmark datasets have shown that our B$^2$Net outperforms other state-of-the-art COD methods. In the future, we will investigate compressing our model into a lightweight model suitable for mobile devices and improving its efficiency in real-time applications.

TABLE III
THE RESULTS OF APPLYING OUR STRATEGY TO OTHER EXISTING MODELS, "-S" MEANS THE IMPROVE MODELS.

| Method | COD10K-Test | | | | Camo-Test | | | | NC4K-Test | | | |
|---|---|---|---|---|---|---|---|---|---|---|---|---|
| | $S_\alpha$ | $E_\phi$ | $F_\beta^\omega$ | M | $S_\alpha$ | $E_\phi$ | $F_\beta^\omega$ | M | $S_\alpha$ | $E_\phi$ | $F_\beta^\omega$ | M |
| BGNet | 0.817 | 0.887 | 0.696 | 0.035 | 0.804 | 0.86 | 0.728 | 0.079 | 0.842 | 0.897 | 0.771 | 0.048 |
| BGNet-S | 0.833 | 0.905 | 0.721 | 0.031 | 0.827 | 0.883 | 0.762 | 0.072 | 0.865 | 0.913 | 0.785 | 0.042 |
| BSANet | 0.831 | 0.901 | 0.722 | 0.033 | 0.816 | 0.871 | 0.751 | 0.069 | 0.851 | 0.907 | 0.788 | 0.044 |
| BSANet-S | 0.848 | 0.922 | 0.742 | 0.027 | 0.833 | 0.895 | 0.781 | 0.061 | 0.878 | 0.922 | 0.801 | 0.038 |

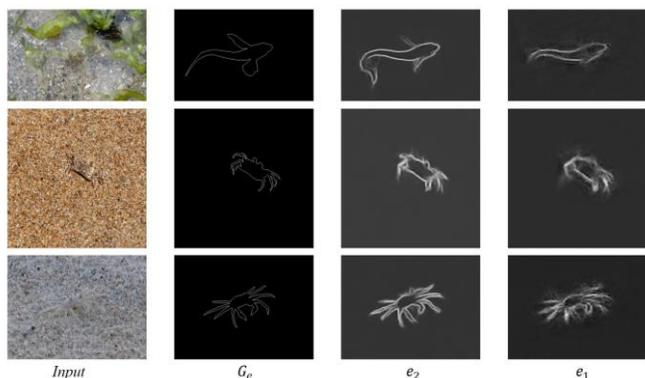

Fig. 7. The visual comparison of the intermediate result e2 and e1, which fully illustrates the effectiveness of our proposed new architecture.